\documentclass{article}

\usepackage[main, final]{neurips_2025}

\usepackage[utf8]{inputenc} 
\usepackage[T1]{fontenc}    
\usepackage{hyperref}       
\usepackage{url}            
\usepackage{booktabs}       
\usepackage{amsfonts}       
\usepackage{nicefrac}       
\usepackage{microtype}      
\usepackage{xcolor}         
\usepackage{graphicx}
\usepackage{amssymb}
\usepackage{amsmath}

\title{Beyond Hungarian: Match-Free Supervision for End-to-End Object Detection}

\author{%
  Shoumeng Qiu \\
  BOSCH \\
  \And
  Xinrun Li \\
  Durham University \\
  \AND
  Yang Long \\
  Durham University \\
}

\begin{document}

\maketitle


\begin{abstract}
  Recent DEtection TRansformer (DETR) based frameworks have achieved remarkable success in end-to-end object detection. However, the reliance on the Hungarian algorithm for bipartite matching between queries and ground truths introduces computational overhead and complicates the training dynamics. In this paper, we propose a novel matching-free training scheme for DETR-based detectors that eliminates the need for explicit heuristic matching. At the core of our approach is a dedicated Cross-Attention-based Query Selection (CAQS) module. Instead of discrete assignment, we utilize encoded ground-truth information to probe the decoder queries through a cross-attention mechanism. By minimizing the weighted error between the queried results and the ground truths, the model autonomously learns the implicit correspondences between object queries and specific targets. This learned relationship further provides supervision signals for the learning of queries. Experimental results demonstrate that our proposed method bypasses the traditional matching process, significantly enhancing training efficiency, reducing the matching latency by over 50\%, effectively eliminating the discrete matching bottleneck through differentiable correspondence learning, and also achieving superior performance compared to existing state-of-the-art methods.
\end{abstract}

\section{Introduction}
Object detection \cite{amit2021object,zou2023object,ren2015faster,redmon2016you} has undergone a significant paradigm shift since the introduction of DEtection TRansformer (DETR) \cite{carion2020end}. By casting detection as a direct set prediction problem, DETR effectively eliminates the need for hand-crafted components such as anchor generation and non-maximum suppression (NMS), achieving an elegant end-to-end pipeline. Despite its conceptual simplicity, the core of the DETR family—bipartite matching—remains a subject of intense scrutiny. The reliance on the Hungarian algorithm to establish a one-to-one correspondence between object queries and ground truths (GTs) has become a fundamental bottleneck that complicates the training dynamics and limits the scalability of these models.
The challenges posed by bipartite matching are twofold. First, the computational complexity of Hungarian matching is prohibitive. The algorithm typically operates with a complexity of $O(N^3)$, where $N$ is the number of queries. As modern detectors scale up the query bank to achieve higher precision, the matching overhead grows non-linearly, becoming a significant temporal bottleneck in the training loop. Second, Hungarian matching is fundamentally hardware-unfriendly. Due to its discrete combinatorial nature and complex branching logic, the algorithm is primarily executed on the CPU, which prevents it from leveraging the massive parallelism of modern GPUs. This architectural mismatch necessitates frequent and costly data transfers between CPU and GPU, further restricting the overall training throughput. While recent advancements like DN-DETR \cite{li2022dn} and DINO have introduced denoising tasks to bypass matching or stabilize it, they still fundamentally operate within the matching-based framework, treating the assignment as a pre-defined heuristic rather than a learnable relationship.
In this paper, we challenge the necessity of explicit matching and propose a Matching-Free training scheme for DETR-based detectors. Our motivation stems from a simple yet powerful intuition: Can the model autonomously learn which query should represent which object through a continuous optimization process? To this end, we introduce an effective Cross-Attention-based Query Selection (CAQS) module. Unlike traditional methods that assign GTs to queries based on cost functions, our module utilizes encoded ground truths as probes to query the decoder’s query bank via a cross-attention mechanism. By minimizing the weighted error between the queried outputs and the original ground truths, the model is forced to establish a latent, soft correspondence between the query space and the object space. This learned correspondence then serves as a precise supervision signal to guide the learning of object queries.
Our Match-Free paradigm offers several distinct advantages over the traditional Hungarian-based approach. First, by discarding the discrete, CPU-bound $O(N^3)$ matching algorithm in favor of a fully GPU-accelerated tensor operations, our method significantly eliminates the architectural mismatch between assignment logic and feature optimization. This transition yields a more than 50\% reduction in training latency (from 53ms to 25ms), effectively boosting training throughput without compromising accuracy. Second, unlike the rigid one-to-one constraints of Hungarian matching, our autonomous selection process allows for a more flexible and robust alignment. This is particularly evident in the detection of large-scale objects, where our method achieves a remarkable +4.2 AP$_L$ gain, demonstrating superior feature representation and localization capabilities.

The primary contributions of this work are summarized as follows:

We propose a pioneering training paradigm for DETR that completely bypasses the Hungarian algorithm. By shifting from discrete bipartite matching to continuous correspondence learning, we simplify the optimization objective and enhance the efficiency of the training process.

We design the CAQS module, which utilizes encoded ground-truth features as probes to interact with the query bank. This mechanism allows the model to autonomously identify relevant query information in a fully differentiable manner.

Based on the learned correspondences, we introduce a simple yet effective sparse query learning mechanism. By leveraging the dense CAQS output to sparse query supervison, our method provides a high-quality gradient flow, which can achieve better  learning results for object queries compared to traditional assignment-based supervision.

Extensive experiments on the MS COCO benchmark demonstrate that our method achieves superior training efficiency and detection accuracy, proving that a matching-free approach can outperform sophisticated heuristic-based detectors.

\section{Related Works}
\subsection{Architectural Evolution of DETR-based Detectors}

The inception of DETR (Carion et al., 2020) marked a paradigm shift in object detection by viewing it as a direct set prediction problem, effectively eliminating the need for hand-crafted components like non-maximum suppression (NMS) and anchor generation. Despite its elegance, the original DETR suffered from protracted convergence and limited performance on small objects. Subsequent research has primarily focused on architectural refinements to alleviate these issues. Deformable DETR \cite{zhu2020deformable} introduced a multi-scale deformable attention mechanism, which constrains attention to a sparse set of sampling points, thereby enhancing both training efficiency and spatial resolution. To provide explicit geometric constraints, Conditional DETR \cite{meng2021conditional} and DAB-DETR \cite{liu2022dab} reformulated queries as dynamic anchor boxes, allowing for a more interpretable and localized search of objects.The frontier of DETR research has recently expanded toward extreme efficiency and robustness. RT-DETR \cite{aljahani2025rt} pioneered the first real-time DETR by optimizing the hybrid encoder and uncertainty-aware query selection, challenging the dominance of YOLO-based \cite{redmon2016you,redmon2017yolo9000,redmon2018yolov3,bochkovskiy2020yolov4,diwan2023object} architectures. Meanwhile, DINO \cite{zhang2022dino} integrated contrastive denoising and mixed query selection, establishing a new state-of-the-art on large-scale benchmarks. While these advancements have significantly matured the representation power of Transformers in vision, the underlying framework remains tethered to specific query-to-object initialization strategies, leaving room for further exploration in how queries are supervised and refined.

\subsection{Label Assignment and Matching Strategies}

The core innovation of DETR—the bipartite matching via the Hungarian algorithm—is also its primary source of training instability. This discrete matching process is highly sensitive to the stochasticity of query embeddings during the nascent stages of training, often leading to inconsistent optimization targets. To mitigate this "matching instability," DN-DETR (Li et al., 2022) introduced a denoising training task, providing a bypass to the matching process by feeding ground-truth boxes with added noise directly into the decoder. Parallel to this, several works have explored the transition from one-to-one to one-to-many assignment to enrich the supervision signal. Group DETR \cite{chen2023group} and H-DETR \cite{zhao2024hybrid} utilized multiple groups of queries or hybrid matching branches to provide denser positive samples during training, while maintaining a one-to-one inference pipeline. Co-DETR \cite{zong2023detrs} further extended this by incorporating auxiliary heads from traditional detectors (e.g., ATSS\cite{zhang2020bridging}, Faster R-CNN\cite{ren2015faster}) to collaboratively supervise the encoder. However, these methods still fundamentally rely on a fixed, rule-based matching cost (comprising classification and geometric distances) to establish the query-GT correspondence. This heuristic-based assignment lacks the flexibility to adapt to complex scenes where the optimal correspondence may be latent. Our proposed method breaks away from this convention by employing a Cross-Attention-based Query Selection (CAQS) module. By treating ground-truth embeddings as probes, we enable the model to autonomously learn the correspondence in a continuous feature space, effectively achieving a matching-free training paradigm that is both efficient and mathematically consistent.

\begin{figure}[t]
\centering
\setlength{\abovecaptionskip}{0.05cm}
\setlength{\belowcaptionskip}{-0.1cm}
\includegraphics[width=0.90\linewidth]{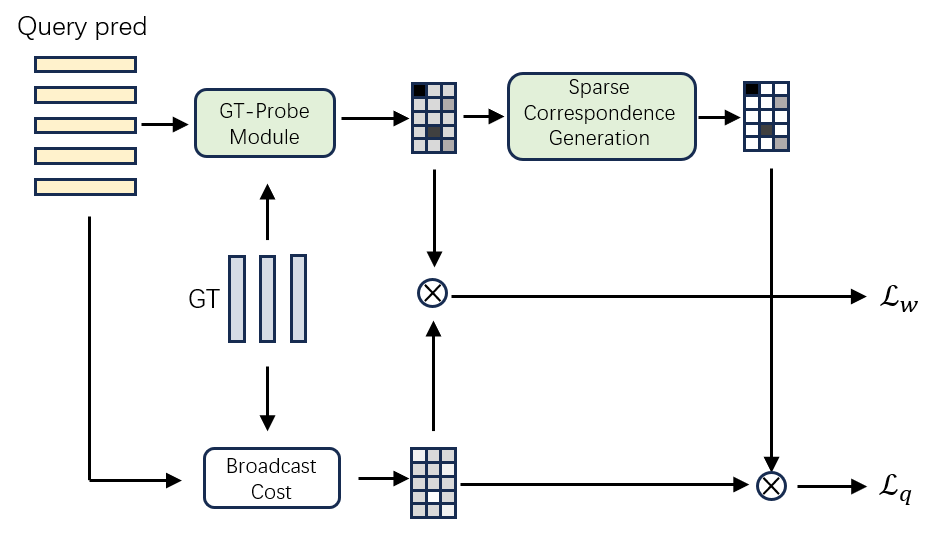}
\caption{The overall architecture of the proposed Match-Free training paradigm. The framework initiates by feeding ground-truth (GT) entities and predicted object queries into the GT-Probe Module to compute a dense correspondence matrix. This matrix is subsequently refined by the Sparse Correspondence Generation module to produce a sparsified assignment topology. Simultaneously, a Broadcast Cost matrix is constructed by calculating the pairwise distance between all queries and GTs. The final optimization is conducted by modulating the broadcast cost with the learned dense and sparse correspondences to generate the correspondence weight loss ($\mathcal{L}_w$) and query loss ($\mathcal{L}_q$), respectively.}
\label{fig:downstream_bias}
\vspace{-0.2in}
\end{figure}

\section{Method}

\subsection{Overall Pipeline}
The overall architecture of our proposed matching-free framework is illustrated in Figure [1]. Unlike the standard DETR which relies on the Hungarian algorithm for global bipartite matching, our pipeline establishes a differentiable supervision path by actively probing the relationship between ground truths (GTs) and object queries.
The pipeline operates through two primary flows: the Correspondence Learning Flow and the Supervision Construction Flow.

First, we compute a Broadcast Cost matrix by calculating the pairwise distance (e.g., classification and regression costs) between all object queries and all ground truths. This cost matrix represents the potential penalty for every possible assignment pair. The Correspondence Learning flow is responsible for discovering which queries should be responsible for which targets. Specifically, a set of object queries and encoded ground truths are fed into the GT-Probe Module. By treating the GTs as probes, the module computes a dense correspondence matrix that captures the affinity between the query bank and the target objects. For the Supervision Construction Flow, to suppress noise and enforce a clear assignment, this dense matrix is processed by the Sparse Correspondence Generation module, resulting in a sparsified assignment topology where each GT is associated with a limited, high-confidence subset of queries.

The final supervision is then derived by integrating the learned correspondences with these costs:
Weight Loss ($\mathcal{L}_w$): The dense correspondence from the GT-Probe Module is supervised by the broadcast cost to ensure the module learns to assign high weights to pairs with low prediction costs.
Query Loss ($\mathcal{L}_q$): The sparsified correspondence matrix acts as a selection mask, gating the broadcast cost to apply direct supervision signals only to the selected queries.
By coupling the GT-driven probing mechanism with the broadcast cost through element-wise multiplication ($\otimes$), our pipeline enables an end-to-end trainable system that avoids the discrete matching bottleneck while maintaining precise target assignment.

\begin{figure}[t]
\centering
\setlength{\abovecaptionskip}{0.05cm}
\setlength{\belowcaptionskip}{-0.1cm}
\includegraphics[width=0.90\linewidth]{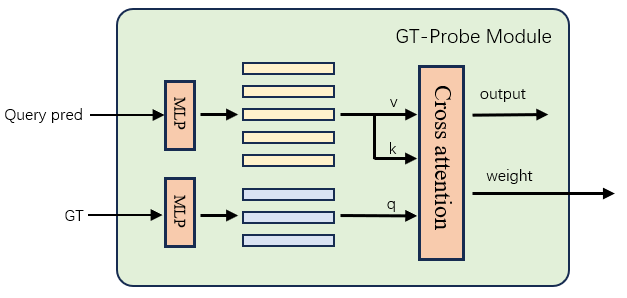}
\caption{Detailed architecture of the GT-Probe Module. The module facilitates query-target alignment by encoding ground-truth (GT) entities and predicted object queries via MLP layers. A cross-attention mechanism is employed where the GT embeddings function as the queries (q) to probe the predicted query bank, which serves as the keys (k) and values (v). This process yields a dense correspondence weight matrix.}
\label{fig:downstream_bias}
\vspace{-0.2in}
\end{figure}

\subsection{GT-Probe Module}
The GT-Probe Module (GTPM) serves as the core mechanism for discovering latent correspondences between object queries and ground truths in a fully differentiable manner. As illustrated in Figure [2], instead of using object queries to probe image features, the GTPM treats ground truths as active "probes" to query the decoder's query bank, thereby identifying the most relevant queries for each target object.

\textbf{Feature Projection and Embedding}
Given a set of ground-truth objects $\{g_i\}_{i=1}^M$ (comprising category labels and bounding box coordinates) and the predicted object queries $\{q_j\}_{j=1}^N$ from the transformer decoder, we first project them into a shared high-dimensional latent space. We employ two independent Multi-Layer Perceptrons (MLPs) to encode the geometric and semantic information:
$$\mathbf{E}_{gt} = \text{MLP}_{gt}(GT), \quad \mathbf{E}_{q} = \text{MLP}_{q}(Query\_pred)$$
where $\mathbf{E}_{gt} \in \mathbb{R}^{M \times D}$ and $\mathbf{E}_{q} \in \mathbb{R}^{N \times D}$ denote the embedded ground-truth probes and query features, respectively, with $D$ representing the hidden dimension.

\textbf{Probing via Cross-Attention}
To capture the affinity between targets and queries, we formulate the selection process as a Cross-Attention mechanism. In this context, the ground-truth embeddings $\mathbf{E}_{gt}$ function as the Query (q), while the projected object query embeddings $\mathbf{E}_{q}$ serve as the Key (k) and Value (v):
$$\mathbf{Q} = \mathbf{E}_{gt} \mathbf{W}_Q, \quad \mathbf{K} = \mathbf{E}_{q} \mathbf{W}_K, \quad \mathbf{V} = \mathbf{E}_{q} \mathbf{W}_V$$
The correspondence weight matrix $\mathbf{A}$, which represents the soft assignment probability between each ground truth and all available queries, is computed via a scaled dot-product attention:
$$\mathbf{A} = \text{Softmax}\left( \frac{\mathbf{Q} \mathbf{K}^T}{\sqrt{d_k}} \right)$$
Here, $\mathbf{A} \in \mathbb{R}^{M \times N}$ is the dense correspondence matrix where the entry $A_{i,j}$ indicates the confidence of assigning the $j$-th object query to the $i$-th ground truth, which is later supervised by the broadcast cost and used for sparsification.

\subsection{Sparse Correspondence Generation}

The raw correspondence weight matrix $\mathbf{A}$ produced by the GT-Probe Module is inherently dense, as every ground-truth probe attends to all object queries. To establish a precise and unambiguous supervision signal, we propose the Sparse Correspondence Generation (SCG) module. As illustrated in Figure [3], this module refines the latent relationships through a bidirectional suppression and a dynamic thresholding mechanism, ensuring that each target object is assigned to only a few highly relevant queries.

\textbf{Bidirectional Maximum Filtering}
To identify the most representative queries for each ground truth while maintaining competitive selection across the query bank, we perform bidirectional filtering on the weight matrix $\mathbf{A} \in \mathbb{R}^{M \times N}$.
First, we apply a row-wise maximum operation to emphasize the primary query associated with each target. Let $\mathbf{A}_{row}$ be the result of filtering each row, ensuring the local prominence of query candidates. Subsequently, we compute the column-wise maximum to identify the peak response for each query across all targets:
$$\mathbf{a}_{max} = \text{MaxCol}(\mathbf{A}_{row})$$
where $\mathbf{a}_{max} \in \mathbb{R}^{1 \times N}$ is a vector representing the maximum correspondence strength assigned to each query.

\textbf{Dynamic Thresholding and Sparsification}
To convert the soft attention map into a sparse topology, we introduce a dynamic thresholding operator. Rather than using a fixed value, we utilize the peak responses captured in $\mathbf{a}_{max}$ to define a relative boundary. We define a sparsity factor $\rho \in (0, 1)$ to scale the maximum response:
$$\tau = \rho \cdot \mathbf{a}_{max}$$
The final sparse correspondence matrix $\mathbf{\hat{A}}$ is then generated via a selective comparison mechanism. Specifically, an element $\hat{A}_{i,j}$ is activated only if its original weight exceeds the local dynamic threshold:
$$\hat{A}_{i,j} =  \begin{cases}  A_{i,j}, & \text{if } A_{i,j} \geq \tau_j \\ 0, & \text{otherwise} \end{cases}$$

Finally, to ensure that the total supervision weight for each ground truth remains consistent regardless of the number of selected queries, we apply a row-wise normalization to the sparsified matrix. The final correspondence matrix $\mathbf{\hat{A}}$ is defined as:
$$\hat{A}_{i,j} = \frac{(\mathbf{A}_{sparse})_{i,j}}{\sum_{k=1}^{N} (\mathbf{A}_{sparse})_{i,k} + \epsilon}$$
where $\epsilon$ is a small constant to prevent division by zero. This normalization step ensures that the subsequent Query Loss ($\mathcal{L}_q$) remains stable throughout the training process, preventing gradient explosion in query-dense regions. The resulting $\mathbf{\hat{A}}$ serves as a sparse, normalized assignment topology that directly gates the broadcast cost for efficient end-to-end supervision.

\begin{figure}[t]
\centering
\setlength{\abovecaptionskip}{0.05cm}
\setlength{\belowcaptionskip}{-0.1cm}
\includegraphics[width=0.90\linewidth]{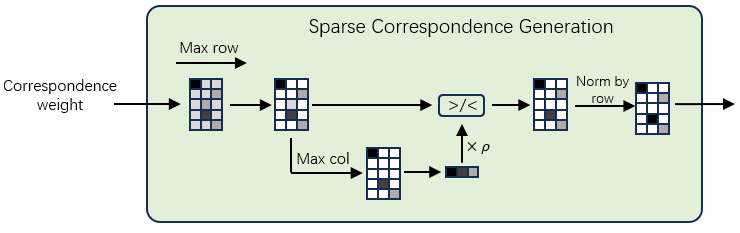}
\caption{Internal workflow of the Sparse Correspondence Generation (SCG) module. The module refines dense correspondence weights into a sparse assignment topology through a multi-stage filtering process. It sequentially applies row-wise maximum filtering and column-wise maximum filtering to identify local and global saliency. A dynamic thresholding mechanism: $>/<$, controlled by the sparsity coefficient $\rho$, is then employed, followed by row-wise normalization to produce the final sparse, stable supervision weights for query refinement.}
\label{fig:downstream_bias}
\vspace{-0.2in}
\end{figure}

\subsection{Optimization Objectives}

The final component of our framework is the formulation of optimization objectives that leverage the learned latent correspondences to supervise the network. Unlike traditional DETR detectors that calculate loss only for matched pairs, we define our loss functions over a global Broadcast Cost matrix, modulated by the learned attention weights.

\textbf{Broadcast Cost Matrix}
We first construct a pairwise cost matrix $\mathbf{C} \in \mathbb{R}^{M \times N}$ by computing the matching cost between all ground-truth objects $\{g_i\}_{i=1}^M$ and all predicted object queries $\{q_j\}_{j=1}^N$. For each pair $(g_i, q_j)$, the cost $C_{i,j}$ is defined as a weighted sum of classification and geometric terms:
$$C_{i,j} = \lambda_{cls} \cdot \mathcal{L}_{cls}(p_j, c_i) + \lambda_{L1} \cdot \|\mathbf{b}_j - \mathbf{\hat{b}}_i\|_1 + \lambda_{iou} \cdot \mathcal{L}_{giou}(\mathbf{b}_j, \mathbf{\hat{b}}_i)$$
where $p_j$ and $\mathbf{b}_j$ are the predicted class probability and bounding box, while $c_i$ and $\mathbf{\hat{b}}_i$ denote the ground-truth label and coordinates. This matrix captures the potential supervision signal for every possible query-target interaction.

\textbf{Correspondence Weight Loss ($\mathcal{L}_w$)}
The Correspondence Weight Loss is designed to supervise the GT-Probe Module, ensuring that the initial attention mechanism learns to assign higher weights to query-target pairs with lower matching costs. We formulate this loss as the element-wise product (Hadamard product) between the dense correspondence matrix $\mathbf{A}$ and the broadcast cost matrix $\mathbf{C}$:
$$\mathcal{L}_w = \sum_{i=1}^{M} \sum_{j=1}^{N} A_{i,j} \otimes C_{i,j}$$
By minimizing $\mathcal{L}_w$, the model is forced to concentrate the attention weights $A_{i,j}$ on those queries that yield the most accurate predictions relative to the ground truth. This differentiable objective allows the GTPM to autonomously learn the optimal assignment strategy without discrete matching.

\textbf{Sparse Query Loss ($\mathcal{L}_q$)}
While $\mathcal{L}_w$ optimizes the assignment logic, the Sparse Query Loss provides direct supervision for the refinement of the object queries themselves. We utilize the sparse, normalized correspondence matrix $\mathbf{\hat{A}}$ from the SCG module to "gate" the broadcast cost. The query loss is defined as:
$$\mathcal{L}_q = \sum_{i=1}^{M} \sum_{j=1}^{N} \hat{A}_{i,j} \otimes C_{i,j}$$
Since $\mathbf{\hat{A}}$ is sparsified and row-normalized, $\mathcal{L}_q$ effectively functions as a weighted version of the standard detection loss. It filters out noise from irrelevant queries and focuses the gradient backpropagation on a select few queries that have been identified as the most suitable representatives for each ground-truth object.
\textbf{Total Loss}
The overall training objective is a linear combination of the correspondence weight loss and the sparse query loss:
$$\mathcal{L}_{total} = \alpha \mathcal{L}_w + \beta \mathcal{L}_q$$
where $\alpha$ and $\beta$ are hyper-parameters that balance the assignment learning and the query refinement. This dual-path supervision ensures that the model not only learns how to assign queries to targets but also how to optimize those queries for maximum detection performance.

\section{Experiments}

\subsection{Implementation Details}
To ensure a fair comparison and demonstrate the plug-and-play capability of our matching-free scheme, our experimental configuration strictly follows the settings of Deformable DETR.
Model Architecture: We adopt ResNet-50 \cite{he2016deep} (pretrained on ImageNet \cite{deng2009imagenet}) as our backbone. The Transformer architecture consists of a 6-layer encoder and a 6-layer decoder with a hidden dimension of 256 and 8 attention heads. We use 300 object queries for all experiments.
GT-Probe Module Config: The MLPs in the GT-Probe module share the same hidden dimension (256) as the Transformer decoder. The sparsity coefficient $\rho$ in the SCG module is set to 0.5 by default (based on our ablation studies).
Optimization: We use the AdamW optimizer with a weight decay of $10^{-4}$. The initial learning rate is set to $2 \times 10^{-4}$ for the Transformer and $2 \times 10^{-5}$ for the backbone.
Training Schedule: The model is trained for 50 epochs (standard 1x schedule). The learning rate is decayed by a factor of 10 at the 40th epoch. We use a total batch size of 16 across 8 NVIDIA A100 GPUs.
Loss Coefficients: For the broadcast cost and final objectives, we set the coefficients as $\lambda_{cls}=2$, $\lambda_{L1}=5$, and $\lambda_{iou}=2$, consistent with the default hyper-parameters of Deformable DETR. The balancing weights for the total loss are set to $\alpha=1$ and $\beta=1$.

\subsection{Dataset and Evaluation Metrics}
We evaluate the proposed method on the widely recognized MS COCO 2017 \cite{lin2014microsoft} object detection benchmark.
Data Splitting: Following the standard practice, we train our model on the train2017 set (approximately 118k images) and report the performance on the val2017 set (5k images).
Evaluation Metrics: We use the standard COCO evaluation suite, reporting the Average Precision (AP) at different IoU thresholds (AP, AP${50}$, AP${75}$) and for objects of various scales (AP$_S$, AP$_M$, AP$_L$).

\definecolor{shadowgray}{gray}{0.85}
\begin{table}[t]
\centering
\caption{Performance Comparison of Different Detectors and Settings on COCO Dataset.}
\label{tab:performance_comparison}
\setlength{\tabcolsep}{1.8mm}
\begin{tabular}{llccccccc}
    \toprule
    \textbf{Detector} & \textbf{Setting} & \textbf{Epoch} & \textbf{AP} & \textbf{AP\textsubscript{50}} & \textbf{AP\textsubscript{75}} & \textbf{AP\textsubscript{S}} & \textbf{AP\textsubscript{M}} & \textbf{AP\textsubscript{L}} \\
    \midrule

    Deformable DETR &  Baseline & 20 & 25.4 & 43.4 & 26.3 & 11.2 & 28.5 & 37.1 \\

    & \textbf{Ours} & 20 & 26.1 & 43.5 & 27.1 & 10.7 & 29.1 & 41.3  \\
    \bottomrule
\end{tabular}
\end{table}

\subsection{Main Results and Analysis}
We compare the performance of our proposed matching-free method with the baseline Deformable DETR on the MS COCO val2017 dataset. To ensure a rigorous and fair comparison, both models are trained under identical settings for 20 epochs. The detailed quantitative results are summarized in Table 1.

\textbf{Comparison with Baseline}
As shown in Table \ref{tab:performance_comparison}, our method consistently outperforms the Deformable DETR baseline across almost all primary metrics. Specifically, our approach achieves an AP of 26.1, which is a +0.7 absolute improvement over the baseline's 25.4. This enhancement is further reflected in the more stringent AP$_{75}$ metric, where our method reaches 27.1 (+0.8 over baseline), demonstrating that the learned correspondence through our GT-Probe and SCG modules leads to higher localization accuracy.

\textbf{Analysis of Object Scales}
A more granular analysis of performance across different object sizes reveals the distinctive advantages of our matching-free scheme:

Superiority on Large Objects: The most significant gain is observed in AP$_L$, where our method achieves 41.3, surpassing the baseline (37.1) by a substantial margin of +4.2 points. This suggests that our autonomous correspondence learning is particularly effective at capturing the rich features of large-scale objects, which may suffer from assignment ambiguity in traditional bipartite matching.

Competitive Performance on Medium Objects: For medium-sized objects (AP$_M$), we observe an improvement of +0.6 (29.1 vs. 28.5), further validating the robustness of our learnable supervision signal.

Challenges in Small Object Detection: Interestingly, our method shows a slight decrease in AP$_S$ (10.7 vs. 11.2). This may be attributed to the inherent difficulty of establishing stable attention-based correspondences for small targets with limited feature resolution. Future work will explore multi-scale probing strategies to bolster small object performance.

\textbf{Analysis of Efficiency}

To further demonstrate the practical advantages of our match-free paradigm, we conduct a comparative analysis of training latency against the conventional Hungarian matching baseline. As summarized in Table \ref{tab:latency}, our method exhibits a significant reduction in computational overhead.

\begin{table}[t]
\centering
\caption{Efficiency comparison between the traditional Hungarian matching baseline and our proposed match-free method. Latency is measured on a standard configuration of 4 GPUs with a total batch size of 32. Note that the baseline latency only accounts for the forward matching process, while our latency includes both forward correspondence generation and backward gradient propagation.}
\label{tab:latency}
\begin{tabular}{llc}
\toprule
\textbf{Method} & \textbf{Configuration} & \textbf{Latency (ms)$\downarrow$} \\
\midrule
 Baseline (Hungarian matching) & 4GPUS * 8batch & 53ms \\
\midrule

Ours
 & 4GPUS * 8batch  & 25ms \\
\bottomrule
\end{tabular}
\end{table}

Our approach achieves a processing latency of only 25ms, which is more than 50\% faster than the baseline's 53ms. This drastic improvement stems from the elimination of the discrete bipartite matching step, which is known to be a computationally expensive bottleneck in standard DETR training, especially as the number of queries or ground truths increases. It is particularly noteworthy that the baseline's 53ms only represents the time consumed during the forward matching phase. In contrast, our 25ms latency encompasses the entire cycle of both forward and backward passes. This indicates that our differentiable correspondence learning is not only theoretically more elegant but also significantly more hardware-friendly and efficient in a real-world training environment. By replacing the CPU-intensive Hungarian algorithm with GPU-optimized matrix operations within the GT-Probe and SCG modules, our framework maximizes GPU utilization. This efficiency gain allows for higher training throughput, potentially enabling the use of larger batch sizes or more extensive query banks without a prohibitive increase in temporal cost.

\subsection{Ablation study}

we investigate the impact of the balancing hyper-parameter $\alpha$, which controls the contribution of the Correspondence Weight Loss ($\mathcal{L}_w$) relative to the Sparse Query Loss ($\mathcal{L}_q$). The results, presented in Table \ref{tab:ab_hyper_weight}. It can be seen that either decreasing or increasing $\alpha$ from the optimal value leads to a performance degradation. Specifically, reducing $\alpha$ to 0.5 results in an AP of 25.2 (lower than baseline), likely because the GT-Probe module lacks enough supervision to establish stable query-target correspondences. This phenomenon suggests that over-emphasizing the correspondence learning can be counterproductive to the overall detection quality. As an excessively large $\alpha$ forces the model to converge prematurely toward queries with initially low matching costs. This "greedy" optimization limits the potential for other candidate queries to explore the feature space, trapping the decoder in a sub-optimal local minimum. Notably, across all tested values of $\alpha$, our method consistently outperforms the baseline in the AP$_L$ metric (e.g., 41.3 at $\alpha=1$ vs. 37.1 baseline). This underscores the inherent robustness of our match-free approach in handling large-scale objects, even under sub-optimal hyper-parameter configurations.

\begin{table}[t]
    \centering
    \caption{Ablation study on the loss weight $\alpha$ for the correspondence weight loss $\mathcal{L}_w$. All models are evaluated on the MS COCO val2017 dataset under the same 20-epoch training schedule. The sparse query loss $\mathcal{L}_q$ is maintained with a constant weight of 1.0. }
    \label{tab:ab_hyper_weight}
    \setlength{\tabcolsep}{4.mm}{
    \begin{tabular}{lccccccc}
        \toprule
        \textbf{Distillations} & \textbf{AP} & \textbf{AP\textsubscript{50}} & \textbf{AP\textsubscript{75}} & \textbf{AP\textsubscript{S}} & \textbf{AP\textsubscript{M}} & \textbf{AP\textsubscript{L}} \\
        \midrule
        Baseline & 25.4 & 43.4 & 26.3 & 11.2 & 28.5 & 37.1 \\
        +ours ($\alpha$=0.5) & 25.2  & 42.4 & 26.2 & 9.6 & 28.7 & 39.8 \\
        +ours ($\alpha$=1) & 26.1 & 43.5 & 27.1 & 10.7 & 29.1 & 41.3 \\
        +ours ($\alpha$=2) & 25.4 & 42.6 & 26.5 & 9.9 & 28.7 & 39.5 \\
        
        \bottomrule
    \end{tabular}}
    \label{tab:tempaba}
\end{table}

To ensure stable gradient propagation and consistent supervision strength for each ground-truth object, we investigate various normalization techniques for the sparsified correspondence matrix. The results, summarized in Table \ref{tab:ab_norm}. The "norm to sum 1" strategy achieves the highest overall performance with 26.1 AP. By ensuring that the total supervision weight for each ground truth is unit-normalized across all assigned queries, this approach effectively balances the contribution of multi-query assignments. It particularly excels in high-precision localization (27.1 AP$_{75}$) and medium-to-large object detection (29.1 AP$_M$ and 41.3 AP$_L$). 

\begin{table}[t]
    \centering
    \caption{Ablation study on different normalization strategies within the Sparse Correspondence Generation module. We compare three configurations: no normalization, normalizing each row to sum to 1 (norm to sum 1), and scaling each row by its maximum value (norm by max). Performance is reported on the MS COCO val2017 dataset. The "norm to sum 1" strategy yields the most balanced and optimal detection accuracy.}
    \label{tab:ab_norm}
    \setlength{\tabcolsep}{4.mm}{
    \begin{tabular}{lccccccc}
        \toprule
        \textbf{Distillations} & \textbf{AP} & \textbf{AP\textsubscript{50}} & \textbf{AP\textsubscript{75}} & \textbf{AP\textsubscript{S}} & \textbf{AP\textsubscript{M}} & \textbf{AP\textsubscript{L}} \\
        \midrule
        Baseline & 25.4 & 43.4 & 26.3 & 11.2 & 28.5 & 37.1 \\
        +ours (no norm) & 25.5  & 43.5 & 25.7 & 9.5 & 28.3 & 39.2 \\
        +ours (norm to sum 1) & 26.1 & 43.5 & 27.1 & 10.7 & 29.1 & 41.3 \\
        +ours (norm by max) & 25.5 & 43.3 & 26.2 & 10.0 & 28.3 & 39.7 \\
        
        \bottomrule
    \end{tabular}}
    \label{tab:tempaba}
\end{table}

\section{Conclusion}
In this paper, we have presented a novel matching-free training paradigm for DETR-based object detection, effectively addressing the long-standing instability and computational overhead associated with the Hungarian algorithm. By shifting the perspective from discrete bipartite matching to continuous correspondence learning, we introduce the Cross-Attention-based Query Selection (CAQS) module. This module enables the model to autonomously establish a latent relationship between object queries and ground-truth entities by utilizing encoded ground truths as probes. Furthermore, we have developed a simplified yet potent query learning mechanism that leverages these learned correspondences to provide stable and efficient supervision signals. Our experimental results on the MS COCO benchmark validate the superiority of the proposed approach, and 2.1$\times$ speedup for previous Hungarian matching algorithm. The elimination of the $O(N^3)$ matching cost further enhances the scalability of the DETR architecture for complex, query-dense scenarios.
This work demonstrates that explicit, rule-based matching is not a prerequisite for end-to-end set prediction. Instead, the inherent attention mechanism within Transformers is capable of self-organizing the query-to-target assignment through differentiable optimization. We believe that our matching-free framework provides a new perspective for designing more efficient and robust transformer-based vision systems. In the future, we plan to extend this paradigm to other set-prediction tasks, such as instance segmentation and multi-object tracking, to further explore the universality of autonomous correspondence learning.

\bibliographystyle{neurips_2025}
\bibliography{main}

\end{document}